%% file: main.tex
\PassOptionsToPackage{table}{xcolor}
\documentclass{article} 
\usepackage{iclr2027_conference,times}

\usepackage{amsmath,amssymb,amsfonts}
\usepackage{booktabs}
\usepackage{graphicx}
\usepackage{multirow}
\usepackage{float}
\usepackage{flafter}
\usepackage{enumitem}
\usepackage{xcolor}
\usepackage{hyperref}
\hypersetup{hidelinks}
\usepackage{url}
\input{math_commands.tex}

\title{\raggedright What Can Latent World Models Know? \\ Physical Information in Multimodal Predictive Representations}

\definecolor{projectlink}{HTML}{2B5FA8}
\author{%
Kaizhen Tan\textsuperscript{1,4,}\thanks{Equal contribution.}\quad
Sizhe Xu\textsuperscript{1,4,*}\quad
Xin Xu\textsuperscript{2}\quad
Siru Tao\textsuperscript{2}\quad
Yixiao Li\textsuperscript{2}\\[2pt]
\bfseries Hanzhe Hong\textsuperscript{2}\quad
Yang Feng\textsuperscript{3}\quad
Heqing Du\textsuperscript{3}\quad
Zhaonan Wang\textsuperscript{4}\\[3pt]
{\normalfont \textsuperscript{1}New York University\quad \textsuperscript{2}Carnegie Mellon University}\\
{\normalfont \textsuperscript{3}Columbia University\quad \textsuperscript{4}NYU Shanghai}\\[2pt]
{\normalfont\textcolor{projectlink}{\href{https://tantansir.github.io/latent-world-model-identifiability/}{Project page: https://tantansir.github.io/latent-world-model-identifiability/}}}
}

\newcommand{\pokeworld}{\textsc{PokeWorld}}
\newcommand{\best}[1]{\textbf{#1}}
\definecolor{tabhead}{gray}{0.92}
\definecolor{tabours}{RGB}{233,240,250}

\iclrfinalcopy

\begin{document}

\maketitle
\fancyhead{}\renewcommand{\headrulewidth}{0pt}
\hypersetup{pdftitle={What Can Latent World Models Know? Physical Information in Multimodal Predictive Representations},
pdfauthor={Kaizhen Tan, Sizhe Xu, Xin Xu, Siru Tao, Yixiao Li, Hanzhe Hong, Yang Feng, Heqing Du, Zhaonan Wang}}

\begin{abstract}
A central premise of latent world models is that predicting the future
encourages representations to internalize the physics of their environment. We
ask which physical quantities are accessible in learned latent states, how
this depends on training, and how those quantities relate to the model's
predictions. We present \pokeworld{}, a simulated environment in which a robot
finger pushes objects whose mass, drag, and contact stiffness vary across
episodes while remaining visually identical. We first measure parameter
recovery from raw observation sequences, then train matched action-conditioned
world models. Prediction targets can strongly shape latent content. For
example, contact stiffness becomes decodable when touch is predicted, while
providing touch only as an input does not. Longer-horizon prediction improves
estimation of position and velocity from learned representations relative to
single-step prediction. Drag is recoverable from raw observations, but has
weak linear readout from the learned latent states. Yet the models' glide
forecasts depend systematically on drag. Experiments on RH20T reproduce the
same input--target dependence on real multimodal robot data. Together, these
results show that observations, prediction targets, and prediction horizons
shape both which physical quantities are accessible in latent states and how
they affect future predictions.
\end{abstract}

\section{Introduction}
\label{sec:intro}

Latent world models learn a state representation by predicting how
observations change under actions \citep{lecun2022path,hafner2025dreamerv3,
assran2025vjepa2}. Joint-embedding predictive architectures (JEPAs) make this
prediction in a learned embedding space instead of pixel space
\citep{assran2023ijepa,balestriero2025lejepa,maes2026lewm}. For a robot, the
value of such a state lies in the physics it captures: how heavy an object is,
how quickly it slows down, and how hard it pushes back on contact. The
prediction loss never names these quantities. Whether a trained latent
contains them, and whether its forecasts depend on them, has to be measured.

Three difficulties make this measurement hard. First, a weak readout can
reflect limited sensor information. Existing physical evaluations include
video benchmarks \citep{garrido2025intuitive,motamed2026physicsiq} and
probe-based readouts from multimodal robot encoders
\citep{tacgen2026,kepler2026}. These measurements alone do not establish what
the observations make recoverable. Second, the training choices
that could matter are rarely varied in isolation; models differ at once in
which sensors they receive, which they predict, and how far ahead they
predict. Third, readout and use are different properties. A probe can extract
a parameter that the model's forecasts ignore, and a forecast can depend on a
parameter that no linear probe extracts.

\begin{figure}[htb]
\centering
\includegraphics[width=\linewidth]{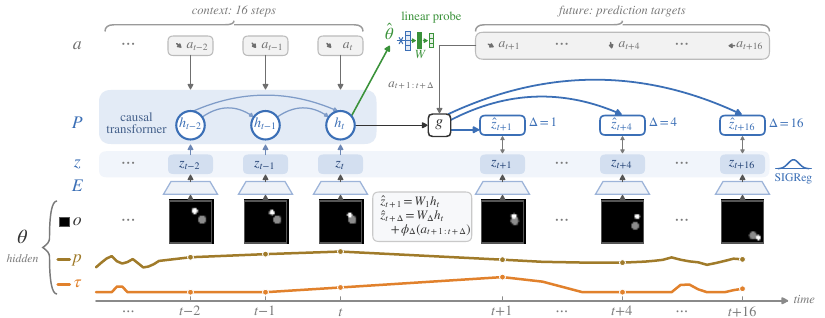}
\caption{\textbf{The world models, one column per time step.} Step $s$ records
a frame $o_s$, proprioception $p_s$, and touch $\tau_s$ (curves: finger speed,
peak force), and $a_s$ is the action applied during it; $\cdots$ marks omitted
steps. $\theta=(m,\gamma,k)$ shapes the observations, not the pixels. The shared
encoder $E$ embeds every step, including the targets; the F variants also feed
it $p_s,\tau_s$. The causal transformer $P$ reads $z_s$ at position $s$,
conditioned on $a_s$, and attends to earlier positions. Heads $g$ predict
embeddings $\Delta\in D=\{1,4,16\}$ steps ahead (for $\Delta>1$ also from
$a_{t+1:t+\Delta}$), each regressed onto the target below it; the X variants
also predict $p,\tau$ at $t{+}1$, $t{+}4$. SIGReg acts on all $z$; a linear
probe $W$ maps the frozen $h_t$ to $\hat\theta$. In training every position
predicts its own future.}
\label{fig:overview}
\end{figure}

We address these difficulties with \pokeworld{}, a simulated pushing
environment in which objects look identical but differ from episode to episode
in mass, drag, and contact stiffness. Because the simulator controls these
parameters, we first measure how well each of them can be recovered from the
raw observation streams, providing empirical recovery references. We then train
action-conditioned JEPA world models that differ in one design choice at a
time: the sensor streams they receive as inputs, the streams they predict as
targets, and the horizons at which they predict (Fig.~\ref{fig:overview}).
Every frozen model is tested twice: whether a linear probe reads a parameter
from its latent state, and whether its forecasts depend on that parameter.
Finally, we test whether the input--target dependence carries over to real
multimodal robot data.

The experiments show that prediction targets shape which physical quantities
are linearly decodable from latent states. Predicting touch makes contact
stiffness decodable, whereas providing touch only as an input does not.
Multi-horizon prediction improves position and velocity readout relative to
next-step prediction. Linear readout and forecast dependence can dissociate
in both directions: weak readout can accompany dependence consistent with
the true dynamics, and strong readout can accompany incorrect dependence.
On real robot data, force remains decodable after training when it is both
supplied as an input and included as a prediction target.

Our contributions are as follows.
\begin{itemize}[leftmargin=*,itemsep=1pt,topsep=2pt]
\item \textbf{\pokeworld{}}, an environment with hidden per-episode physics
and multimodal sensor streams, with empirical recovery references from the
observations and tests of how a model's forecasts depend on each
parameter (Secs.~\ref{sec:pokeworld} and~\ref{sec:method}).
\item \textbf{A controlled analysis} showing that prediction targets and
horizons affect physical-parameter and motion readout, that
linear readout and forecast dependence dissociate in both directions, and that
the objective of the prediction head limits the predicted effect of a new
action (Sec.~\ref{sec:exp}).
\item \textbf{Evidence on real multimodal robot data} that the same
input--target dependence holds outside simulation (Sec.~\ref{sec:real}).
\end{itemize}

\section{Related Work}
\label{sec:related}

\paragraph{Latent world models.} I-JEPA and V-JEPA predict masked targets in
embedding space \citep{assran2023ijepa,assran2025vjepa2}, and
action-conditioned variants plan with predicted embeddings
\citep{assran2025vjepa2,zhou2025dinowm}. LeJEPA replaces teacher networks with
SIGReg \citep{balestriero2025lejepa}, and LeWorldModel trains such a model end
to end from pixels \citep{maes2026lewm}; DreamerV3 combines reconstruction with
latent dynamics \citep{hafner2025dreamerv3}. \citet{terver2025physplan} study
which design choices make JEPA world models plan well. Concurrent work restores
state that a LeWorldModel latent omits with auxiliary targets
\citep{zhu2026spectral,yan2026psgjepa,zeng2026phylatent,liu2026grounded}.
\citet{liu2026tcsigreg} apply SIGReg to temporally centered residuals to
preserve time-varying state information. JEPA-x shows that a state target can
raise decodability without improving forecasts \citep{wen2026jepax};
Semigroup-JEPA conditions
latent dynamics on given physical parameters \citep{liu2026semigroup}. We
study hidden per-episode parameters and whether forecasts use them.

\paragraph{Probing physical knowledge.} Intuitive-physics benchmarks evaluate
video models \citep{garrido2025intuitive,motamed2026physicsiq}, and
CALIPER shows that frozen encoders decode mass- and friction-dependent outcomes
from pixel displacement \citep{mehta2026caliper}; \citet{joseph2026interpreting}
locate and intervene on physical quantities inside video world models. Probe
results depend on probe capacity \citep{hewitt2019control,belinkov2022probing}.
System identification asks whether a parameter is recoverable under a given
observation and excitation process \citep{ljung1999system}, and recent theory
states identifiability conditions for JEPA representations, action-conditioned
transitions, and physical dynamics from video \citep{klindt2026lejepa,
zhang2026identifiability,wang2026physicsvideo,dong2026closure}; other work
analyzes JEPA risk and SIGReg and the link between prediction error and control
\citep{jepatheory2026,sigregfe2026,controltheory2026}. We pair each readout
with a recovery reference from the same observations and with the model's own
forecasts.

\paragraph{Multimodal and physical estimation.} Sparsh learns tactile
representations \citep{higuera2025sparsh}, TacGen improves contact-property
decoding with tactile features \citep{tacgen2026}, TouchWorld couples tactile
prediction with action generation \citep{touchworld2026}, Kepler-Encoder fuses
vision, proprioception, and force--torque \citep{kepler2026}, and HPT
pretrains across embodiments \citep{hpt2024}. Video-based physical estimation
recovers object parameters from observed interaction \citep{wu2015galileo,
wu2016physics101,xu2019densephysnet}. Our factorial separates receiving a
modality as input from predicting it as a target.

\section{\pokeworld{}}
\label{sec:pokeworld}

\begin{figure}[htb]
\centering
\includegraphics[width=\linewidth]{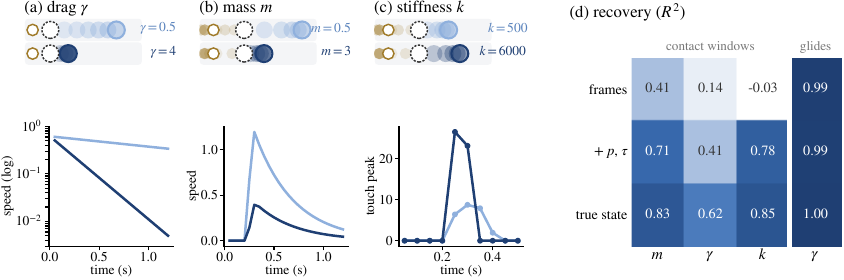}
\caption{\textbf{Hidden physics in \pokeworld{}.} (a)--(c)~Matched episodes
start from the same state and receive the same actions; only one parameter
differs (light: low value, dark: high value). Top: the object every $0.2$\,s,
so that spacing shows speed (dashed: start; gold: finger). Bottom: the stream
in which the parameter shows. (d)~Supervised recovery $R^2$ from 16-step
windows of the frames (as the image-derived
object trajectory), of the frames with proprioception and touch, and of the
true object state, on contact windows and on pure glides. The supplementary
material animates the three pairs.}
\label{fig:pokeworld}
\end{figure}

\pokeworld{} hides physical parameters from every single frame and reveals
them only through how the scene evolves (Fig.~\ref{fig:pokeworld}).

\paragraph{Hidden physics.} A force-controlled finger pushes one object in a
two-dimensional arena. Each episode draws the object's mass $m$, drag
coefficient $\gamma$, and contact stiffness $k$, none of which affects the
rendering. The object's velocity $\mathbf v$ follows
\begin{equation}
m\,\dot{\mathbf v}=\mathbf f_c-\gamma m\,\mathbf v+\mathbf f_w,\qquad
\mathbf f_c=k\,d^{3/2}\,\mathbf n,
\label{eq:dyn}
\end{equation}
where $d$ is the finger--object overlap, $\mathbf n$ the contact normal, and
$\mathbf f_w$ the wall force; light contact damping preserves impact
transients (App.~\ref{app:env}). Each parameter leaves its trace in a
different part of the motion (Fig.~\ref{fig:pokeworld}a--c). Drag sets the
decay of a gliding object, $|\mathbf v(t)|=|\mathbf v(0)|\,e^{-\gamma t}$,
independently of its mass. Mass sets the velocity change that a contact
impulse produces. Stiffness sets the height and duration of the contact-force
peak and changes the resulting motion only slightly.

\paragraph{Sensor streams.} At every control step $t$ the environment emits a
$64{\times}64$ grayscale frame $o_t$, the finger's proprioception $p_t$
(position and velocity), touch $\tau_t$ (the mean, peak, and end-of-step
reaction force within the step, the fraction of the step in contact, and a
contact flag), and the finger force $a_t$ commanded during the step. Sub-step
touch statistics keep the impact peak that averaging over a step would erase. A
mixed behavior policy produces sustained pushes, impacts, and free glides
(App.~\ref{app:env}).

\paragraph{Empirical recovery reference.} We measure recovery of a parameter
$\theta_j$ from a set of streams by fitting a supervised estimator and
evaluating it on held-out episodes. A two-layer GRU $f_j$ reads 16-step windows
$x$ of the
streams and actions, with the frames summarized by the image-derived object
trajectory, and its held-out coefficient of determination
\begin{equation}
\rho^{\mathrm{rec}}_j=1-\frac{\sum_i\big(\theta_{ij}-f_j(x_i)\big)^2}
{\sum_i\big(\theta_{ij}-\bar\theta_j\big)^2}
\label{eq:rec}
\end{equation}
is an empirical recovery reference for $\theta_j$: performance achieved by
this GRU, not an information-theoretic upper bound. Recovery depends on the
streams and context (Fig.~\ref{fig:pokeworld}d). Stiffness recovery improves
with touch, and mass recovery improves with touch and proprioception. Drag
recovery is lower on contact-dominated windows, even from the true object
state, than on free glides, where frames alone give nearly exact recovery.
Where recovery is high but latent readout is low, the tested probes access
less of the parameter information available in the observations.

\section{World Models and Measurements}
\label{sec:method}

The accessibility of physical quantities in a world model may depend on
what it is trained to predict. We therefore study action-conditioned JEPA world
models that differ in three sets: the input streams $I$, the target streams
$U$, and the prediction horizons $D$ (Fig.~\ref{fig:overview}).

\paragraph{Model.} An encoder $E$ embeds each step, and a causal transformer
$P$ with AdaLN action conditioning summarizes the history into the predictor
state $h_t$:
\begin{equation}
z_t=E\big(o_t,\{u_t\}_{u\in I}\big),\qquad h_t=P\big(z_{\le t},a_{\le t}\big),
\qquad I\subseteq\{p,\tau\}.
\label{eq:model}
\end{equation}
Latent heads predict the embedding $\Delta$ steps ahead from $h_t$ and the
intervening actions, $\hat z_{t+\Delta}=g_\Delta(h_t,a_{t+1:t+\Delta})$, where
the next-step head reads $h_t$ alone; sensor heads predict the raw future
values $\hat u_{t+\Delta}$ of every stream in a target set $U\subseteq\{p,\tau\}$.
The model minimizes
\begin{equation}
\mathcal L=\sum_{\Delta\in D}w_\Delta\big\|\hat z_{t+\Delta}-z_{t+\Delta}\big\|^2
+\sum_{u\in U}\sum_{\Delta\in D_U}w^{u}_{\Delta}\big\|\hat u_{t+\Delta}-u_{t+\Delta}\big\|^2
+\lambda\,\mathrm{SIGReg}(z),
\label{eq:loss}
\end{equation}
where SIGReg keeps the embeddings close to an isotropic Gaussian and prevents
collapse without a target network \citep{balestriero2025lejepa,maes2026lewm}.
Unless stated otherwise, $D=\{1,4,16\}$ and $D_U=\{1,4\}$ (App.~\ref{app:training}).

\paragraph{Variants.} Model names encode $I$ and $U$. V receives vision and
predicts only embeddings ($I=U=\emptyset$). F adds proprioception and touch as
inputs ($I=\{p,\tau\}$), and X adds them as targets ($U=\{p,\tau\}$; X$_p$ and
X$_t$ add one of them). VF thus receives every stream without predicting any,
and VFX receives and predicts all of them. All variants share data,
architecture, and compute.

\paragraph{Readout.} For a frozen model, a ridge probe maps $h_t$ at the end of
a 16-step context to $\theta_j$. Its held-out coefficient of determination,
computed as in Eq.~\ref{eq:rec} on the same windows, is the readout
$\rho^{\mathrm{read}}_j$. The gap
$\rho^{\mathrm{rec}}_j-\rho^{\mathrm{read}}_j$ compares recovery from the
observations with linear readout from the latent; it does not by itself
establish that the latent has discarded the parameter information.

\paragraph{Forecast dependence.} A model can use a parameter that no linear
probe extracts, or ignore one that a probe extracts. For a forecast $\hat y$
of a physical quantity, such as the object's displacement or the peak contact
force, we regress $\hat y$ and the simulator's outcome $y$ on $\theta_j$, with
the context state and the other parameters as covariates, and report the slope
ratio
\begin{equation}
S=\beta_{\hat y\mid\theta_j}\,/\,\beta_{y\mid\theta_j},
\label{eq:slope}
\end{equation}
which is $1$ when the forecast depends on $\theta_j$ as the physics does and
$0$ when it ignores $\theta_j$. The continuation dependence $S^{\mathrm{nat}}$
uses the recorded future actions. The response dependence $S^{\Delta}$ applies
two prescribed action sequences $A_1$ and $A_0$ from the same state and
measures the dependence of the action effect $\Delta y=y(A_1)-y(A_0)$, with
true outcomes from replaying the simulator. Its error
\begin{equation}
E^{\Delta}=\textstyle\sum_i\big(\Delta\hat y_i-\Delta y_i\big)^2\big/\sum_i\Delta y_i^{2}
\label{eq:effect}
\end{equation}
is $0$ for exact effects and $1$ for a model that predicts no effect.

\paragraph{Intervention models.} To compare linear readout with forecast
dependence, we train three models that receive the object's position and
velocity next to the images and predict with an interaction head, an MLP over
$h_t$ and the action window. The baseline uses VFX inputs and targets; the
coordinate-target model adds auxiliary targets in which each parameter enters
linearly (App.~\ref{app:training}), and the reference model receives the true
parameters. Each test starts where the parameter matters: a push against a
resting object for mass, a push timed to hit in the fourth frame for
stiffness, and the release of a pushed object for drag.

\paragraph{Frozen-head diagnostic.} To separate representation content from
the objective used to train the forecast head, we freeze each predictor and
train new heads $H$ on the push test. A head receives $h_t$, the current
object and finger state $s_t$, an action window $A_b$, and a parameter vector
$c_\alpha=\hat\theta+\alpha(\theta-\hat\theta)$, where $\hat\theta$ is a ridge
estimate from $h_t$; $\alpha=0$ adds no parameter information beyond $h_t$ and
$\alpha=1$ supplies the true parameters. With $\hat d_b$ the displacement that
a fixed position probe decodes from $H(h_t,s_t,A_b,c_\alpha)$, the same head
is trained on the same samples with either
\begin{equation}
\mathcal L_{\mathrm{lat}}=\sum_{b\in\{0,1\}}\big\|H(h_t,s_t,A_b,c_\alpha)-z^{(b)}_{t+16}\big\|^2
\quad\text{or}\quad
\mathcal L_{\mathrm{eff}}=\big\|(\hat d_1-\hat d_0)-(d_1-d_0)\big\|^2,
\label{eq:heads}
\end{equation}
where $z^{(b)}_{t+16}$ and $d_b$ are the embedding and displacement that the
simulator produces under $A_b$.

\section{Experiments}
\label{sec:exp}

\paragraph{Setup.} Models train on $4{,}000$ \pokeworld{} episodes of $64$
steps. Probes are fitted and evaluated on separate episodes, and tests with
prescribed actions use freshly simulated contexts stratified over the tested
parameter (Apps.~\ref{app:training} and~\ref{app:measure}). We report means
over three training seeds; the appendix lists every seed.

\begin{table}[htb]
\caption{\textbf{Prediction targets affect physical-parameter readout.}
Readout $\rho^{\mathrm{read}}$ (mean $\pm$ s.d.\ over three seeds)
from the frozen predictor state on contact windows. The last two rows are the
recovery references of Fig.~\ref{fig:pokeworld}d on the same windows.}
\label{tab:factorial}
\begin{center}
\small
\begin{tabular}{lllccc}
\toprule
\rowcolor{tabhead} Model & Inputs & Targets & $\log m$ & $\gamma$ & $\log k$ \\
\midrule
V & vision & latent & $0.13\pm0.01$ & $0.07\pm0.01$ & $-0.03\pm0.01$ \\
VF & vision, P, T & latent & $0.21\pm0.00$ & $0.14\pm0.01$ & $-0.01\pm0.00$ \\
VX$_p$ & vision & latent, P & $0.17\pm0.01$ & $0.09\pm0.01$ & $-0.01\pm0.00$ \\
VX$_t$ & vision & latent, T & $0.25\pm0.01$ & $0.12\pm0.01$ & $0.43\pm0.01$ \\
VX & vision & latent, P, T & $0.26\pm0.00$ & $0.13\pm0.01$ & $0.47\pm0.01$ \\
\rowcolor{tabours} VFX & vision, P, T & latent, P, T & \best{$0.29\pm0.01$} & $0.12\pm0.01$ & \best{$0.50\pm0.01$} \\
\midrule
\multicolumn{3}{l}{Recovery from the VFX inputs} & $0.71$ & $0.41$ & $0.78$ \\
\multicolumn{3}{l}{Recovery from the true object state} & $0.83$ & $0.63$ & $0.85$ \\
\bottomrule
\end{tabular}
\end{center}
\end{table}

\paragraph{Prediction targets shape parameter readout.}
Table~\ref{tab:factorial} varies inputs and targets. Touch as a target, not as
an input, makes stiffness decodable: VF receives the full touch stream and
reads stiffness at chance, whereas VX$_t$, which predicts touch without
receiving it, reads it at $0.43$, and VFX at $0.50$. Predicting
proprioception instead leaves stiffness at chance, and removing the contact
frames from the touch loss removes the stiffness readout of VFX
(App.~\ref{app:more}). Mass follows the same order more weakly, and every
model stays far below its recovery reference. Drag readout stays at about a third
of its reference or less in these models; direct parameter supervision raises
it above that reference (App.~\ref{app:more}).

\begin{table}[htb]
\caption{\textbf{Multi-horizon prediction improves position and velocity readout.}
Within each observation and regularization setting, vision-only V models
differ only in their latent prediction heads (three seeds). Object position
is read from the frame embedding and velocity
from the predictor state; $a_{t+1}$ is the action that produces the next frame.
The three-head and multi-horizon rows share a summed loss weight of $2$.}
\label{tab:horizon}
\begin{center}
\small
\resizebox{\linewidth}{!}{\begin{tabular}{lcccccc}
\toprule
\rowcolor{tabhead} & \multicolumn{2}{c}{grayscale, $\lambda=0.1$} & \multicolumn{2}{c}{grayscale, $\lambda=0.02$} & \multicolumn{2}{c}{frame + difference, $\lambda=0.02$} \\
\rowcolor{tabhead} Latent prediction heads & position & velocity & position & velocity & position & velocity \\
\midrule
one next-step head & $0.05\pm0.01$ & $0.11\pm0.02$ & $0.85\pm0.03$ & $0.30\pm0.01$ & $0.67\pm0.06$ & $0.40\pm0.01$ \\
one next-step head, given $a_{t+1}$ & $0.05\pm0.01$ & $0.14\pm0.02$ & $0.82\pm0.03$ & $0.33\pm0.01$ & $0.68\pm0.04$ & $0.40\pm0.01$ \\
three next-step heads, given $a_{t+1}$ & $0.34\pm0.23$ & $0.18\pm0.08$ & $0.91\pm0.02$ & $0.39\pm0.00$ & $0.85\pm0.05$ & $0.43\pm0.01$ \\
heads at $\Delta\in\{1,4,16\}$ & $0.90\pm0.04$ & $0.35\pm0.04$ & $0.99\pm0.00$ & $0.64\pm0.01$ & $0.95\pm0.01$ & $0.78\pm0.01$ \\
\bottomrule
\end{tabular}}
\end{center}
\end{table}

\paragraph{Longer horizons improve motion readout.}
Table~\ref{tab:horizon} compares vision-only models within each observation
and regularization setting. A single next-step head yields weaker velocity
readout and, under the stronger regularizer, weak position readout
($R^2=0.05$), while adding heads at $\Delta=4$ and $16$ improves both
(position $0.90$). Giving the next-step head the next action has little effect,
and three next-step heads with the same total loss weight also yield lower
velocity readout than the multi-horizon model. Longer horizons improve motion
readout across the tested settings.

\begin{figure}[htb]
\centering
\includegraphics[width=\linewidth]{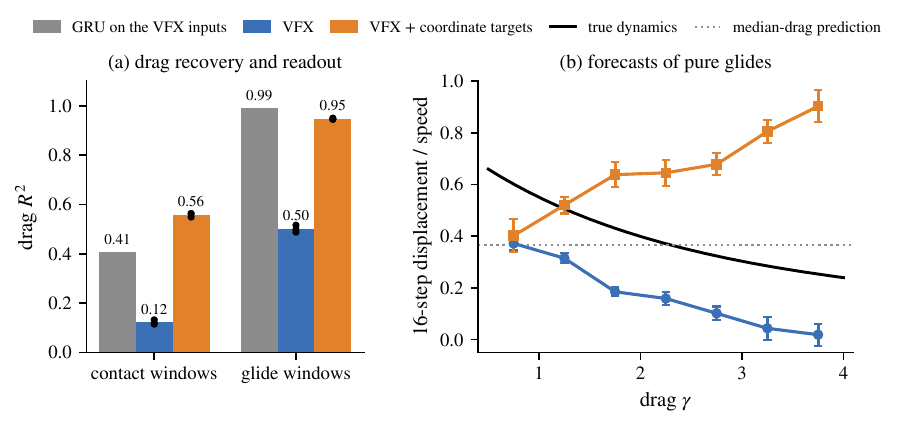}
\caption{\textbf{Drag is recoverable, read on glides, and used by glide
forecasts.} (a) Drag $R^2$ on contact windows and on pure glides for a GRU on
the VFX inputs (recovery reference) and for ridge readout from the frozen
predictor state of VFX and of the coordinate-target model defined in
Sec.~\ref{sec:method}, which also receives the object state (points: seeds).
(b) Predicted 16-step displacement divided by the speed at the end of the
context, per drag bin (mean $\pm$ s.d.\ over three seeds), with the true
dynamics and a prediction that assumes the median drag.}
\label{fig:glide}
\end{figure}

\paragraph{Drag is read on glides, and forecasts follow it.} Drag shows only in
the deceleration of a gliding object, which contact windows rarely contain. On
pure glides the supervised GRU recovers drag almost exactly, and the VFX latent
reads it far better than on contact windows, though still well below its
reference (Fig.~\ref{fig:glide}a). Its 16-step forecasts shorten with drag at
the rate of the true dynamics (slope ratio $1.04$), with a bias toward short
displacements (Fig.~\ref{fig:glide}b). Readability does not guarantee correct
use: the coordinate-target model of Table~\ref{tab:three} reads drag on the
same glides almost perfectly, yet its forecast displacement grows with drag.

\begin{table}[htb]
\caption{\textbf{Readout versus forecast dependence.} Three models with
object-state input and interaction heads: the VFX baseline, the same model
with coordinate targets, and a model given the true parameters. Readout is
measured at the start of each parameter's test; $S$ is the slope ratio of a
forecast (Eq.~\ref{eq:slope}; $1$: as the physics) and $E^{\Delta}$ the error
of the predicted action effect (Eq.~\ref{eq:effect}). Means over three seeds
(App.~\ref{app:three}).}
\label{tab:three}
\begin{center}
\small
\begin{tabular}{llccc}
\toprule
\rowcolor{tabhead} Parameter & Measurement & Baseline & Coordinate targets & True parameters \\
\midrule
mass $m$ & readout & $0.10$ & $0.48$ & $0.84$ \\
 & $S^{\mathrm{nat}}$, recorded continuation & $0.19$ & $0.36$ & $0.98$ \\
 & $S^{\Delta}$, push vs.\ zero force & $0.05$ & $0.23$ & $0.56$ \\
 & $E^{\Delta}$, push vs.\ zero force & $0.68$ & $0.72$ & $0.65$ \\
\midrule
stiffness $k$ & readout & $0.67$ & $0.83$ & $0.93$ \\
 & $S^{\mathrm{nat}}$, recorded continuation & $0.68$ & $0.74$ & $0.97$ \\
 & $S^{\Delta}$, push vs.\ retreat & $0.62$ & $0.72$ & $0.89$ \\
 & $E^{\Delta}$, push vs.\ retreat & $0.33$ & $0.31$ & $0.28$ \\
\midrule
drag $\gamma$ & readout & $0.06$ & $0.41$ & $0.70$ \\
 & $S$, release & $0.07$ & $0.12$ & $0.55$ \\
\bottomrule
\end{tabular}
\end{center}
\end{table}

\paragraph{Readable is not the same as used.} Table~\ref{tab:three} compares
linear readout and forecast dependence for the intervention models defined in
Sec.~\ref{sec:method}. Coordinate targets
make every parameter more readable at the start of its test, but the forecasts
respond far less. The mass dependence of the predicted push effect reaches a
quarter of the simulator's, less than half of what the true-parameter model
reaches, and its error does not fall; the drag dependence of the release
forecast stays near zero. Stiffness is the exception, because peak force is
itself a prediction target.

\begin{figure}[htb]
\centering
\includegraphics[width=0.93\linewidth]{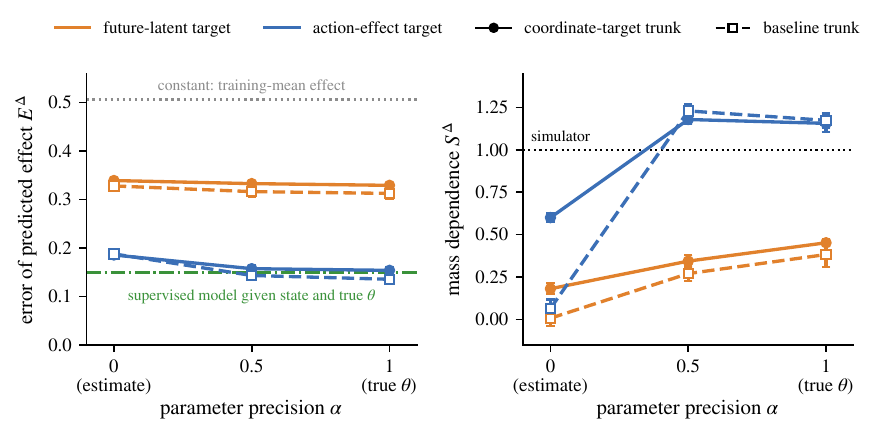}
\caption{\textbf{The objective of the head limits the predicted action
effect.} New heads on frozen predictors predict the effect of a 16-step push
from the predictor state, the current object and finger state, the actions,
and $c_\alpha=\hat\theta+\alpha(\theta-\hat\theta)$, trained on future
latents ($\mathcal L_{\mathrm{lat}}$) or on the action effect
($\mathcal L_{\mathrm{eff}}$). Left: error of the predicted effect. Right: its
mass dependence. Error bars: s.d.\ over three trunk seeds.}
\label{fig:effect}
\end{figure}

\paragraph{The objective of the head limits the response.} On the coordinate-target predictor,
$\mathcal L_{\mathrm{eff}}$ nearly halves the error of
$\mathcal L_{\mathrm{lat}}$ at $\alpha=0$ ($E^{\Delta}=0.19$
versus $0.34$) and approaches a supervised model given the true parameters,
whereas $\mathcal L_{\mathrm{lat}}$ barely improves even when it receives the
true parameters (Fig.~\ref{fig:effect}). The objective of the prediction head,
more than the content of the representation, limits how accurately the model
predicts what a new action will do.

\section{Real Multimodal Robot Data}
\label{sec:real}

\paragraph{Setup.} RH20T records robot manipulation with synchronized
cameras, proprioception, and wrist force--torque \citep{fang2024rh20t}. We
train V, VX, VF, and VFX on the episodes of one robot and evaluate on tasks
held out from training (App.~\ref{app:real}). The F variants add the
end-effector pose and force--torque to a $96{\times}96$ grayscale view, and the
X variants predict both. Because the recorded action is the next end-effector
pose, we evaluate force, which the actions do not reveal: current force read
from the frame embedding $z_t$, and force $0.4$\,s ahead read from the
predictor state $h_t$.

\begin{table}[htb]
\caption{\textbf{Real robot data (RH20T, Flexiv): predicting force preserves
its readout when force is included in the inputs.} Held-out-task $R^2$ (mean $\pm$ s.d.\ over three seeds) of
current force read from the frame embedding and of force $0.4$\,s ahead read
from the predictor state. The untrained model is the VFX network at
initialization. For future force, a GRU trained on the raw sensor history
reaches $0.46$ and persistence $0.31$.}
\label{tab:rh20t}
\begin{center}
\small
\begin{tabular}{lllcc}
\toprule
\rowcolor{tabhead} Model & Inputs & Targets & Current force & Future force \\
\midrule
V & image & latent & $-0.28\pm0.13$ & $-0.15\pm0.07$ \\
VX & image & latent, pose, force & $-0.21\pm0.02$ & $-0.09\pm0.06$ \\
VF & image, pose, force & latent & $-0.16\pm0.08$ & $-0.10\pm0.05$ \\
\rowcolor{tabours} VFX & image, pose, force & latent, pose, force & \best{$0.88\pm0.03$} & \best{$0.47\pm0.00$} \\
\midrule
untrained & image, pose, force & none & $0.91$ & $0.38$ \\
\bottomrule
\end{tabular}
\end{center}
\end{table}

\begin{figure}[htb]
\centering
\includegraphics[width=0.87\linewidth]{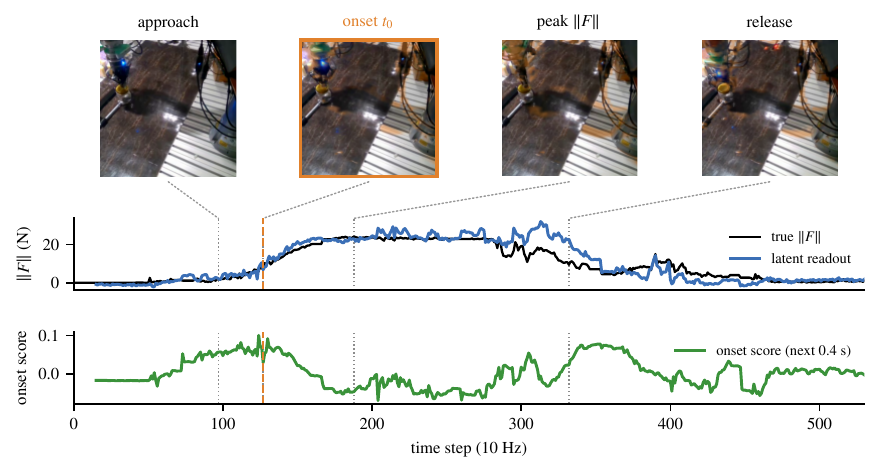}\\[4pt]
\includegraphics[width=0.87\linewidth]{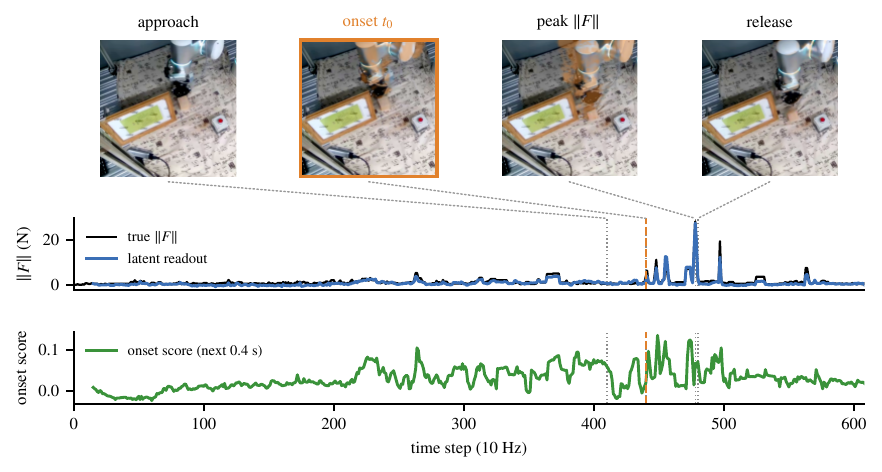}
\caption{\textbf{Force read from the frozen VFX latent on held-out episodes}
of two robots (top: KUKA; bottom: Flexiv). Each block shows four camera
frames, measured force magnitude (black), force read from the latent (blue),
and the readout score for contact within the next $0.4$\,s (green). The orange
dashed line marks the measured contact onset. The model receives
$96{\times}96$ grayscale images rather than the color crops shown.}
\label{fig:qual}
\end{figure}

\paragraph{Results.} Table~\ref{tab:rh20t} reproduces the input--target
dependence. Current-force readout is high for both the untrained fused-input
encoder and VFX. Training VF with the same inputs on visual targets alone
reduces readout $R^2$ below zero. Future force separates the models in the same way:
VF falls below the untrained model, while VFX matches a GRU trained directly
on the raw sensor history. On a second robot, too, only VFX reads force on
held-out tasks (App.~\ref{app:real}). Figure~\ref{fig:qual} follows force on
two held-out episodes: the readout from the frozen VFX latent tracks a
sustained press and repeated impulses, and its contact score rises before the
measured onset.

\section{Discussion}
\label{sec:limitations}

\paragraph{Why prediction targets affect readout.} Under the squared error of
Eq.~\ref{eq:loss}, the optimal head outputs the conditional mean of each
target given $h_t$, so a parameter lowers the loss only if it moves this mean.
Stiffness barely changes the object's motion (Fig.~\ref{fig:pokeworld}c), so
future embeddings hardly depend on it, whereas the touch peak does; stiffness
readout improves when touch is a target and falls when contact frames are
removed from the touch loss. Likewise, velocity barely
changes the next frame but changes the frame sixteen steps later, and over a
contact window the future depends far more on the push than on drag. Providing
a stream as input alone does not ensure that its information remains
accessible to the tested probes.

\paragraph{Implications for training world models.} Three practices follow.
A quantity whose readout matters can be included in the prediction targets.
Predicting at several horizons improves motion readout in our experiments.
A world model meant for control should be tested
on the effects of actions it has not seen, because neither readout nor
continuation accuracy predicted how well our models answered a new action.

\paragraph{Limitations.} \pokeworld{} is two-dimensional with three hidden
parameters, and our models have about five million parameters; larger visual
world models may yield higher readout by default. The models in Table~\ref{tab:three}
and Fig.~\ref{fig:effect} receive the object state, and their coordinate
targets are computed from simulator state. RH20T provides no parameter labels,
so there we test force rather than hidden parameters. World models with
explicit episode-level latent variables, which infer the parameters as part of
prediction, are the natural next test.

\section{Conclusion}
\label{sec:conclusion}

Predicting the future does not by itself ensure high physical-parameter
readout. In \pokeworld{}, prediction targets shape parameter readout, and
longer horizons improve position and velocity readout. Drag readout improves
on glides, where forecasts follow drag at the rate of the true dynamics.
Readability and
use remain different properties: a model can read drag almost perfectly and
still forecast glides that lengthen with drag, and training the head on the
action effect instead of on future latents nearly halves the error of a
predicted push. The same input--target dependence holds on real robot data.
These results support evaluating physical information through both linear
readout and forecast dependence, under specified observations, targets, and
prediction horizons.

\subsection*{AI use statement}
In this work, we used generative AI tools to help develop the conceptual
framework of the study, propose and refine hypotheses, design and give feedback
on experiments, implement methods (the simulator, the models, and the
evaluations), clean and reformat the robot dataset, and interpret results. We
have not used generative AI tools to generate data sets, to formulate
mathematical claims, or for translation: every reported number comes from our
simulator and from training and evaluating our models, and the paper states no
theorems, so proofs and qualitative or thematic data analysis are not
applicable. Additionally, we used generative AI tools to create and edit
software code and scientific figures, suggest experimental parameters, search
for, identify, and summarize literature, suggest the structure of the paper,
draft and edit parts of the paper, and format references. We have reviewed all
AI-assisted work: we tested the code, checked every reported number against the
outputs of our experiments, checked interpretations against the results, and
verified bibliographic information against the original publications. We take
responsibility for the final content of this work, including text, claims or
artifacts produced with the aid of generative AI.

\subsubsection*{Reproducibility statement}
The environment, models, and evaluations run on a single GPU. The appendix
lists every configuration, and code, data generators, and checkpoints will be
released.

\label{endofmain}
\bibliography{refs_generated,iclr2027_conference}
\bibliographystyle{iclr2027_conference}

\appendix
\raggedbottom

\section{\pokeworld{} Details}
\label{app:env}

\paragraph{Dynamics.} Every episode samples $m\in[0.5,3]$ and
$k\in[500,6000]$ log-uniformly and $\gamma\in[0.5,4]$ uniformly. Bodies follow
semi-implicit Euler integration at $\mathrm{d}t{=}0.05$ with 20 substeps. The finger (mass $1$, radius $0.06$) is
driven by a commanded force $F_{\max}a_t$ with $a_t\in[-1,1]^2$. The object
(radius $0.09$) feels the drag force $-\gamma m v$, so its speed decays at rate
$\gamma$ regardless of mass. Finger--object contact follows the Hertzian
penalty $F_c=k\,d^{3/2}$ with normal damping at $0.25$ of critical, where $d$
is the overlap; underdamping preserves the impact peak through which touch
carries stiffness. The action $a_t$ is the command applied during step $t$, so
the state $s_t$ is the result of $a_t$.

\paragraph{Sensors.} Vision is an anti-aliased $64{\times}64$ grayscale frame
(object intensity about $0.55$, finger $1.0$) together with its temporal
difference. Proprioception is the finger's position and velocity. Touch
reports sub-step statistics of the reaction force on the finger in each control
step: mean force (2-d), peak magnitude, end-of-step force (2-d), fraction of
the step in contact, and a contact flag. Force channels carry noise with
standard deviation $0.02$. Sub-step statistics are necessary because averaging
the force over a control step removes the impact peak.

\paragraph{Matched episodes.} The pairs in Fig.~\ref{fig:pokeworld}a--c are
replayed open loop from one initial state with one action sequence. The drag
pair launches the object at speed $0.62$ with the finger at rest ($m=1.2$,
$k=2000$); the mass and stiffness pairs strike a resting object with five steps
of full force followed by a retreat ($\gamma=2.5$; $k=2000$ for the mass pair,
$m=1.2$ for the stiffness pair).

\paragraph{Behavior policy and data.} Episodes of 64 steps mix pursuit,
strike-and-retreat, and Ornstein--Uhlenbeck drift with random switching, and a
fraction of episodes launch the object with a random velocity. The training
set has $4{,}000$ episodes; probes are fitted on $1{,}600$ and evaluated on
$800$ independent episodes. Tests with prescribed actions draw contexts from
$48{,}000$ freshly simulated episodes per test.

\section{Model and Training Details}
\label{app:training}

\paragraph{Architecture.} The per-frame encoder is a four-stage CNN (channels
$32/64/128/256$, stride 2) over the image channels, with MLPs for
proprioception and touch, fused by a linear projection with BatchNorm into a
128-dimensional embedding; following \citet{maes2026lewm}, there is no trailing
LayerNorm. The predictor is a four-layer causal transformer (width 192, eight
heads, dropout 0.1) with zero-initialized AdaLN action conditioning. The
next-step head is a linear projection of $h_t$ with BatchNorm. Heads at
$\Delta\in\{4,16\}$ are additive in $h_t$ and an MLP embedding of the
flattened action window, $\hat z_{t+\Delta}=W_h h_t+W_a\phi(a_{t+1:t+\Delta})+b$.
Cross-modal heads predict raw touch and proprioception at $t{+}1$ from $h_t$
and at $t{+}4$ with $a_{t+1:t+4}$. Loss weights are $1$ for the next-step
latent, $0.5$ for the other latent heads and the next-step cross-modal heads,
and $0.25$ for the four-step cross-modal heads; contact frames carry weight 4
in the touch loss.

\paragraph{Optimization.} AdamW with learning rate $3{\times}10^{-4}$, cosine
decay, 500 warmup steps, and weight decay $0.05$; batches of 128 windows of 24
frames; $20{,}000$ steps in bf16. SIGReg has weight $\lambda=0.02$ unless
stated otherwise and uses 1024 random directions and 17 integration knots,
resampled at every step. A model trains in about seven
minutes on one H200 GPU.

\paragraph{Models of Table~\ref{tab:three}.} The object's position and
velocity enter the encoder next to the image features. Interaction heads
replace the additive heads with a two-layer MLP (width 256) over $h_t$ and the
action embedding. The coordinate targets are predicted from the predictor
state with loss weight $0.5$ each: for drag, the four-step change
$\log|v_{t+4}|-\log|v_t|=-4\gamma\,\mathrm{d}t$ on contact-free windows in
which the object keeps moving; for stiffness,
$\log|f_{t+1}|-1.5\log d_{t+1}=\log k$ on contact frames; for mass,
$\log|\bar f_{t+1}|-\log a_{\parallel}=\log m$ on contact frames away from the
walls, where $a_\parallel$ is the object's acceleration along the force plus
the drag term $\gamma\bar v$. The true-parameter model receives
$(\log m,\gamma,\log k)$ as additional conditioning of every predictor layer.

\paragraph{Control variants.} The reconstruction control replaces the latent
prediction losses with a deconvolutional pixel decoder (four stride-2
upconvolutions from a $4{\times}4$ map, channels $256/128/64/32$) and removes
SIGReg; cross-modal losses stay active. The parameter-supervision control adds
a two-layer head from $h_t$ to the whitened $(\log m,\gamma,\log k)$ with
weight $0.5$. The Gaussian-NLL control adds variance heads to every latent
prediction.

\section{Measurement Details}
\label{app:measure}

\paragraph{Windows and probes.} Probe windows are 16-step contexts starting at
steps $0,8,\dots,40$ of probe episodes. Contact windows contain at least two
contact frames ($3{,}228$ test windows); moving windows have a mean object
speed above $0.15$. Ridge probes standardize the features and select the
penalty on a held-out part of the fit episodes; MLP probes are early-stopped
two-layer MLPs. Object position is read from the frame embedding $z_t$ and
object velocity from $h_t$ on moving windows.

\paragraph{Recovery references.} The GRU references (two layers, width 96)
read 16-step windows and stop early on $15\%$ of the fit episodes. Image-derived
trajectories use the object's intensity band after removing the finger disk,
which is located from the image, with central-difference velocities.
Table~\ref{tab:matched} lists the references on the probe windows. Contact
windows rarely contain a free glide, so for drag we also use the pure-glide
windows of Fig.~\ref{fig:glide} (one window per episode, stratified over
$\gamma$) with five-fold cross-validation over episodes. There, the GRU
recovers drag at $0.99$ from the frames alone, at $0.99$ from the VFX inputs,
and at $1.00$ from the true object state (Fig.~\ref{fig:pokeworld}d), and the
structured estimator $\hat\gamma=-\operatorname{median}_t\log(|v_{t+1}|/|v_t|)/\mathrm{d}t$
reaches $0.96$ from image-derived speeds and $1.00$ from simulator speeds.

\begin{table}[H]
\caption{\textbf{Recovery references.} Test $R^2$ for $\log m$ / $\gamma$ /
$\log k$ by input set and window class; GRU, with ridge regression on the
flattened window in parentheses. V: image-derived object position and
velocity; P: proprioception; T: touch; A: actions.}
\label{tab:matched}
\begin{center}
\small
\begin{tabular}{llccc}
\toprule
\rowcolor{tabhead} Windows & Inputs & $\log m$ & $\gamma$ & $\log k$ \\
\midrule
contact & V+A & $0.41$ ($-0.01$) & $0.14$ ($0.00$) & $-0.03$ ($0.00$) \\
contact & V+P+T+A & $0.71$ ($0.03$) & $0.41$ ($-0.01$) & $0.78$ ($0.20$) \\
contact & object state+P+T+A & $0.83$ ($0.03$) & $0.63$ ($-0.01$) & $0.85$ ($0.20$) \\
all & V+A & $0.24$ ($-0.01$) & $0.15$ ($0.00$) & $-0.05$ ($0.00$) \\
all & V+P+T+A & $0.45$ ($0.02$) & $0.41$ ($-0.01$) & $0.49$ ($0.13$) \\
all & object state+P+T+A & $0.52$ ($0.02$) & $0.56$ ($-0.01$) & $0.51$ ($0.13$) \\
\bottomrule
\end{tabular}
\end{center}
\end{table}

\paragraph{Tests with prescribed actions.} Mass contexts end with the object
slower than $0.10$ and out of contact for two frames, contain at least two
contact frames, and have the finger $0.16$--$0.45$ from the object, away from
the walls. The push applies $0.9\,\mathrm{unit}(\mathrm{object}-\mathrm{finger})$
for 16 steps; the alternative applies zero commanded force, and the finger then
moves under the simulator dynamics. Stiffness contexts end with the finger
$0.02$--$0.10$ from the object after at least two contact frames; push and
retreat apply $\pm0.9$ along the same direction for four steps, and we keep
starts where the push first makes contact in the fourth frame and the retreat
never does. Drag contexts end with the object in contact for two frames and
moving faster than $0.2$ with room ahead; the finger then withdraws for 16
steps. Contexts are stratified over the tested parameter
with at most $300$ in each of six bins, giving $1{,}788$ mass, $1{,}455$
stiffness, and $1{,}800$ drag contexts. Slopes control for the finger--object distance, the finger speed,
the finger and object velocities along the action direction, the direction, and
the other two parameters (for stiffness: gap, normal and tangential relative
velocity, direction, $\log m$, $\gamma$). Mass and drag forecasts decode the
$\Delta{=}16$ prediction with a linear position probe; stiffness forecasts use
the peak-force channel of the $\Delta{=}4$ touch head. Readout at the start of a
test is an episode-grouped ridge probe on the same contexts.

\paragraph{Frozen-head diagnostic.} Mass contexts are drawn from independent
episode sets for training ($96{,}000$ episodes, $5{,}426$ starts), validation
($1{,}500$ starts), and testing (the $1{,}788$ starts above). Each start stores
$h_t$, the current object and finger state, and the replayed outcomes of both
branches: 16-step displacements $d_1,d_0$ and the frozen encoder's embeddings
$z_1,z_0$ of the true frame 16. $\hat\theta$ is a ridge regression from $h_t$
to the standardized parameters, cross-fitted over five episode folds on the
training starts; its test $R^2$ for $(\log m,\gamma,\log k)$ is
$0.51/0.44/0.62$ on the coordinate-target predictor and $0.12/0.03/0.31$ on the
baseline. The head is a two-layer MLP (width 256) from
$(h_t,s_t,\phi(A),c_\alpha)$ to an embedding decoded by the fixed position
probe. The latent target is the squared error to $z_1$ and $z_0$; the
action-effect target is the squared error of $\hat d_1-\hat d_0$ to $d_1-d_0$.
Both use the same samples, initialization, optimizer (AdamW, $3{\times}10^{-4}$,
4{,}000 steps), and validation-based selection. The supervised reference is an
MLP on $(s_t,\theta,\text{direction})$ trained on the same starts. Per-trunk
values are reported in App.~\ref{app:three}.

\section{Additional \pokeworld{} Results}
\label{app:more}

\paragraph{Nonlinear probes.} Early-stopped MLP probes on the same contact
windows reproduce the ridge results of Table~\ref{tab:factorial}
(Table~\ref{tab:mlp}).

\begin{table}[H]
\caption{\textbf{MLP probes.} Early-stopped two-layer MLP readout ($R^2$, mean
$\pm$ s.d.\ over three seeds) from the frozen predictor state on contact
windows.}
\label{tab:mlp}
\begin{center}
\small
\begin{tabular}{lccc}
\toprule
\rowcolor{tabhead} Model & $\log m$ & $\gamma$ & $\log k$ \\
\midrule
V & $0.00\pm0.00$ & $-0.01\pm0.00$ & $-0.00\pm0.00$ \\
VF & $0.00\pm0.00$ & $-0.00\pm0.00$ & $-0.00\pm0.00$ \\
VX$_p$ & $-0.00\pm0.00$ & $-0.01\pm0.00$ & $-0.00\pm0.00$ \\
VX$_t$ & $0.20\pm0.01$ & $0.04\pm0.01$ & $0.42\pm0.00$ \\
VX & $0.22\pm0.01$ & $0.06\pm0.01$ & $0.45\pm0.00$ \\
VFX & $0.26\pm0.02$ & $0.05\pm0.01$ & $0.50\pm0.01$ \\
\bottomrule
\end{tabular}
\end{center}
\end{table}

\paragraph{Objective controls.} Keeping the VFX predictor, data, and
probes fixed, we change only the objective (Table~\ref{tab:controls}).
Supervising the parameters directly raises every parameter readout, most for
stiffness and mass; pixel reconstruction in place of latent prediction leaves
drag readout low, and stiffness remains decodable with touch forecasting.

\begin{table}[H]
\caption{\textbf{Objective controls} (one seed). Ridge readout from the
predictor state on contact windows; position from the frame embedding;
velocity on moving windows.}
\label{tab:controls}
\begin{center}
\small
\begin{tabular}{lccccc}
\toprule
\rowcolor{tabhead} Objective & $\log m$ & $\gamma$ & $\log k$ & position & velocity \\
\midrule
VFX & $0.30$ & $0.11$ & $0.49$ & $0.98$ & $0.78$ \\
VFX + parameter supervision & $0.73$ & $0.46$ & $0.92$ & $0.97$ & $0.76$ \\
VFX, Gaussian NLL & $0.30$ & $0.12$ & $0.46$ & $1.00$ & $0.88$ \\
VFX, pixel reconstruction & $0.18$ & $0.04$ & $0.57$ & $0.95$ & $0.65$ \\
V, pixel reconstruction & $0.16$ & $0.08$ & $-0.01$ & $0.69$ & $0.50$ \\
\bottomrule
\end{tabular}
\end{center}
\end{table}

\paragraph{Regularizer weight.} A lighter SIGReg weight sharpens every
readout of the VFX model (Table~\ref{tab:lambda}).

\begin{table}[H]
\caption{\textbf{SIGReg weight} (VFX, one seed). Ridge readout on contact
windows; velocity on moving windows.}
\label{tab:lambda}
\begin{center}
\small
\begin{tabular}{lccccc}
\toprule
\rowcolor{tabhead} $\lambda$ & $0.3$ & $0.1$ & $0.02$ & $0.01$ & $0.005$ \\
\midrule
$\log k$ & $0.13$ & $0.30$ & $0.49$ & $0.51$ & $0.57$ \\
$\log m$ & $0.14$ & $0.19$ & $0.30$ & $0.32$ & $0.32$ \\
$\gamma$ & $0.00$ & $0.05$ & $0.11$ & $0.13$ & $0.12$ \\
velocity & $0.54$ & $0.63$ & $0.78$ & $0.78$ & $0.80$ \\
\bottomrule
\end{tabular}
\end{center}
\end{table}

\paragraph{Touch-loss weight on contact frames.} Stiffness readout depends on
how strongly the touch loss weighs contact frames, while mass and drag barely
move (Table~\ref{tab:touchw}).

\begin{table}[H]
\caption{\textbf{Contact-frame weight of the touch loss} (VFX, one seed; the
default is 4). Ridge readout on contact windows.}
\label{tab:touchw}
\begin{center}
\small
\begin{tabular}{lcccc}
\toprule
\rowcolor{tabhead} weight & $4$ & $1$ & $0.25$ & $0$ \\
\midrule
$\log k$ & $0.49$ & $0.31$ & $0.10$ & $0.00$ \\
$\log m$ & $0.30$ & $0.28$ & $0.26$ & $0.22$ \\
$\gamma$ & $0.11$ & $0.13$ & $0.14$ & $0.15$ \\
\bottomrule
\end{tabular}
\end{center}
\end{table}

\section{Per-Seed Results for Readout versus Forecast Dependence}
\label{app:three}

The additive heads of Table~\ref{tab:factorial} give an action effect
$W_a[\phi(A_1)-\phi(A_0)]$ that does not depend on the history, and their mass
dependence of the push effect is $0.01$, $0.00$, and $-0.01$ in the three
seeds. The tables below list every seed, the direct-supervision and
self-conditioning variants, and the frozen-head results of
Fig.~\ref{fig:effect}.

\begin{table}[H]
\caption{\textbf{Mass and drag tests, per seed.} Every slash separates the
three representation seeds. Columns: mass push-branch
slope ratio, mass action-effect ratio $S^{\Delta}$ and relative error
$E^{\Delta}$, mass continuation ratio $S^{\mathrm{nat}}$, and drag
release-branch slope ratio. Models: additive = VFX with the additive head and no state input; interaction, no state = interaction head without state input; baseline, coordinate, oracle = the three models of Table~\ref{tab:three} (oracle: true parameters); + self-cond.\ = coordinate targets with the model's own coordinate estimates fed to the predictor; direct sup.\ = parameter-supervision heads, no state input.}
\label{tab:threeseeds}
\begin{center}
\scriptsize
\setlength{\tabcolsep}{3pt}
\begin{tabular}{lccccc}
\toprule
\rowcolor{tabhead} model & $m$ push & $m$ $S^{\Delta}$ & $m$ $E^{\Delta}$ & $m$ $S^{\mathrm{nat}}$ & $\gamma$ release \\
\midrule
additive & 0.06/0.03/-0.01 & 0.01/-0.00/-0.01 & 0.54/0.52/0.54 & 0.03/0.00/-0.04 & -0.01/-0.03/-0.06 \\
interaction, no state & 0.03/0.08/0.09 & 0.07/0.09/0.04 & 0.85/0.80/0.90 & 0.14/0.14/0.12 & -0.01/-0.00/-0.03 \\
baseline & 0.06/0.11/0.12 & 0.04/0.00/0.11 & 0.64/0.77/0.64 & 0.16/0.19/0.23 & 0.06/0.10/0.05 \\
coordinate & 0.37/0.29/0.41 & 0.22/0.13/0.35 & 0.73/0.75/0.69 & 0.37/0.36/0.35 & 0.10/0.15/0.12 \\
+ self-cond. & 0.34/0.29/0.37 & 0.21/0.19/0.22 & 0.66/0.63/0.76 & 0.35/0.33/0.40 & 0.08/0.11/0.11 \\
direct sup. & 0.34/0.43/0.29 & 0.27/0.39/0.28 & 1.01/1.01/0.89 & 0.26/0.33/0.32 & -0.00/0.00/0.01 \\
oracle & 1.12/1.23/1.04 & 0.58/0.71/0.39 & 0.63/0.69/0.62 & 0.98/0.97/0.98 & 0.54/0.55/0.56 \\
\bottomrule
\end{tabular}
\end{center}
\end{table}

\begin{table}[H]
\caption{\textbf{Stiffness test, per seed} (slashes separate seeds). Push-branch slope ratio, action-effect ratio $S^{\Delta}$ and
relative error $E^{\Delta}$ (push minus retreat), continuation ratio
$S^{\mathrm{nat}}$, and the stiffness readout at the impact start.}
\label{tab:kseeds}
\begin{center}
\scriptsize
\begin{tabular}{lccccc}
\toprule
\rowcolor{tabhead} model & $k$ push & $k$ $S^{\Delta}$ & $k$ $E^{\Delta}$ & $k$ $S^{\mathrm{nat}}$ & $k$ readout at start \\
\midrule
additive & 0.39/0.39/0.40 & 0.00/0.00/0.00 & 1.00/1.00/1.00 & 0.53/0.53/0.54 & 0.37/0.41/0.44 \\
interaction, no state & 0.59/0.59/0.56 & 0.60/0.61/0.54 & 0.39/0.38/0.38 & 0.63/0.66/0.61 & 0.62/0.61/0.61 \\
baseline & 0.62/0.62/0.62 & 0.63/0.61/0.61 & 0.35/0.32/0.30 & 0.67/0.68/0.68 & 0.65/0.67/0.69 \\
coordinate & 0.69/0.63/0.70 & 0.75/0.69/0.73 & 0.31/0.32/0.30 & 0.77/0.69/0.77 & 0.81/0.83/0.85 \\
+ self-cond. & 0.76/0.66/0.72 & 0.78/0.68/0.74 & 0.25/0.25/0.24 & 0.79/0.66/0.74 & 0.86/0.83/0.85 \\
direct sup. & 0.69/0.70/0.67 & 0.70/0.72/0.65 & 0.39/0.36/0.37 & 0.74/0.75/0.73 & 0.86/0.87/0.87 \\
oracle & 0.87/0.88/0.88 & 0.89/0.89/0.88 & 0.30/0.29/0.25 & 0.96/0.97/0.97 & 0.94/0.93/0.93 \\
\bottomrule
\end{tabular}
\end{center}
\end{table}

\begin{table}[H]
\caption{\textbf{Readouts of the same checkpoints} (three-seed means). Readout at the
intervention start is the episode-grouped ridge $R^2$ on the mass-push
contexts ($m$), the release contexts ($\gamma$), and the impact contexts
($k$); contact and glide columns are the ridge probes of Table~\ref{tab:factorial}
on the standard probe windows. The oracle is given the parameters.}
\label{tab:threeread}
\begin{center}
\scriptsize
\begin{tabular}{lccccccc}
\toprule
\rowcolor{tabhead} model & start $m$ & start $\gamma$ & start $k$ & contact $m$ & contact $\gamma$ & contact $k$ & glide $\gamma$ \\
\midrule
additive & 0.03 & 0.05 & 0.41 & 0.29 & 0.12 & 0.50 & 0.13 \\
interaction, no state & 0.07 & 0.05 & 0.61 & 0.38 & 0.13 & 0.64 & 0.08 \\
baseline & 0.10 & 0.06 & 0.67 & 0.45 & 0.14 & 0.65 & 0.08 \\
coordinate & 0.48 & 0.41 & 0.83 & 0.76 & 0.56 & 0.82 & 0.82 \\
+ self-cond. & 0.50 & 0.40 & 0.84 & 0.78 & 0.55 & 0.83 & 0.82 \\
direct sup. & 0.43 & 0.25 & 0.87 & 0.75 & 0.47 & 0.92 & 0.43 \\
oracle & 0.84 & 0.70 & 0.93 & 0.76 & 0.49 & 0.85 & 0.43 \\
\bottomrule
\end{tabular}
\end{center}
\end{table}

Heads trained on future latents with the actions alone (Table~\ref{tab:ladder}) reproduce the pattern of Fig.~\ref{fig:effect}: true parameters raise the mass dependence of the push effect but not its accuracy, and an estimate extracted from the same representation adds no more than a random projection of it.
Table~\ref{tab:effect} gives the results for the heads in Fig.~\ref{fig:effect},
which also receive the current object and finger state.

\begin{table}[H]
\caption{\textbf{New heads on frozen predictors, trained on future latents.}
$H(h,A)$ uses the predictor state and the actions; $H(h,A,\hat\theta)$ adds a
cross-fitted ridge estimate of the parameters from $h$; $H(h,A,Ph)$ adds a
fixed random three-dimensional projection of $h$ instead; $H(h,A,\theta)$ adds
the true parameters. Mass push-branch slope ratio, mass dependence $S^{\Delta}$
and error $E^{\Delta}$ of the push effect, and drag release slope; means over
three seeds.}
\label{tab:ladder}
\begin{center}
\small
\begin{tabular}{llcccc}
\toprule
\rowcolor{tabhead} Frozen predictor & Head & $m$ push & $m$ $S^{\Delta}$ & $m$ $E^{\Delta}$ & $\gamma$ release \\
\midrule
baseline & $H(h,A)$ & $0.12$ & $0.02$ & $0.68$ & $0.10$ \\
 & $H(h,A,\hat\theta)$ & $0.14$ & $0.12$ & $0.67$ & $0.09$ \\
 & $H(h,A,Ph)$ & $0.12$ & $0.09$ & $0.67$ & $0.09$ \\
 & $H(h,A,\theta)$ & $0.68$ & $0.65$ & $0.66$ & $0.07$ \\
\midrule
coordinate targets & $H(h,A)$ & $0.39$ & $0.25$ & $0.66$ & $0.13$ \\
 & $H(h,A,\hat\theta)$ & $0.43$ & $0.32$ & $0.71$ & $0.12$ \\
 & $H(h,A,Ph)$ & $0.41$ & $0.27$ & $0.67$ & $0.12$ \\
 & $H(h,A,\theta)$ & $0.91$ & $0.76$ & $0.68$ & $0.12$ \\
\bottomrule
\end{tabular}
\end{center}
\end{table}

\begin{table}[H]
\caption{\textbf{Training target versus parameter precision on the mass action effect.} New heads on frozen trunks
receive the trunk state, the current object and finger state, the action window, and
$c_\alpha=\hat\theta+\alpha(\theta-\hat\theta)$, and are trained with either the latent target or the
action-effect target on the same intervention samples; $\alpha=0$ adds no test-time parameter information.
Columns: action-effect slope ratio $S^{\Delta}$, relative error $E^{\Delta}$, RMSE (arena units), and skill
relative to the training-mean effect, on $1{,}788$ held-out test starts; mean $\pm$ s.d.\ over three trunk seeds with the
head-training seed fixed. The constant and the true-parameter control do not depend on the trunk (fixed).}
\label{tab:effect}
\begin{center}
\small
\begin{tabular}{lcccc}
\toprule
\rowcolor{tabhead} head & $S^{\Delta}$ & $E^{\Delta}$ & RMSE & skill \\
\midrule
\rowcolor{tabhead} \multicolumn{5}{l}{coordinate-target trunk (3 seeds)} \\
\midrule
constant, training-mean effect (fixed) & -0.00 & 0.51 & 0.108 & 0.00 \\
supervised control, $(s_t,\theta,\text{dir})$ (fixed) & 1.07 & 0.15 & 0.058 & 0.71 \\
supervised control, $(s_t,\hat\theta,\text{dir})$ & 0.56$\pm$0.00 & 0.17$\pm$0.00 & 0.062$\pm$0.001 & 0.67$\pm$0.01 \\
latent target, $\alpha=0$ & 0.18$\pm$0.03 & 0.34$\pm$0.00 & 0.088$\pm$0.001 & 0.33$\pm$0.01 \\
latent target, $\alpha=0.5$ & 0.34$\pm$0.03 & 0.33$\pm$0.00 & 0.087$\pm$0.000 & 0.34$\pm$0.01 \\
latent target, $\alpha=1$ & 0.45$\pm$0.02 & 0.33$\pm$0.00 & 0.087$\pm$0.000 & 0.35$\pm$0.01 \\
action-effect target, $\alpha=0$ & 0.60$\pm$0.03 & 0.19$\pm$0.01 & 0.065$\pm$0.001 & 0.63$\pm$0.01 \\
action-effect target, $\alpha=0.5$ & 1.18$\pm$0.03 & 0.16$\pm$0.01 & 0.060$\pm$0.001 & 0.69$\pm$0.02 \\
action-effect target, $\alpha=1$ & 1.16$\pm$0.05 & 0.15$\pm$0.01 & 0.059$\pm$0.001 & 0.70$\pm$0.01 \\
\midrule
\rowcolor{tabhead} \multicolumn{5}{l}{baseline trunk (3 seeds)} \\
\midrule
supervised control, $(s_t,\hat\theta,\text{dir})$ & 0.05$\pm$0.06 & 0.19$\pm$0.01 & 0.065$\pm$0.001 & 0.63$\pm$0.01 \\
latent target, $\alpha=0$ & 0.01$\pm$0.05 & 0.33$\pm$0.01 & 0.087$\pm$0.001 & 0.35$\pm$0.01 \\
latent target, $\alpha=0.5$ & 0.27$\pm$0.04 & 0.32$\pm$0.01 & 0.085$\pm$0.001 & 0.38$\pm$0.02 \\
latent target, $\alpha=1$ & 0.38$\pm$0.07 & 0.31$\pm$0.01 & 0.085$\pm$0.002 & 0.38$\pm$0.02 \\
action-effect target, $\alpha=0$ & 0.06$\pm$0.05 & 0.19$\pm$0.01 & 0.066$\pm$0.001 & 0.63$\pm$0.01 \\
action-effect target, $\alpha=0.5$ & 1.23$\pm$0.04 & 0.14$\pm$0.00 & 0.057$\pm$0.000 & 0.72$\pm$0.00 \\
action-effect target, $\alpha=1$ & 1.17$\pm$0.04 & 0.14$\pm$0.00 & 0.056$\pm$0.000 & 0.73$\pm$0.00 \\
\bottomrule
\end{tabular}
\end{center}
\end{table}

\section{RH20T Details}
\label{app:real}

\paragraph{Data and preprocessing.} We use the Flexiv configuration of
RH20T ($4{,}258$ episodes; $2{,}947$ training episodes from 112 tasks; $479$
evaluation episodes from 12 held-out tasks) and the KUKA configuration ($896$
episodes, 56 tasks) at $10$\,Hz. Models receive a $96{\times}96$ grayscale view
from one camera; the F variants add the end-effector pose and the six-axis
force--torque signal. Streams are clipped to the $0.1$--$99.9$ percentiles of
the training episodes and whitened. The $800$-episode subset is sampled per
task and has its own clipping and whitening statistics; the future-force label,
the persistence prediction, and the evaluation windows are identical at both
sizes. The recorded action at step $t$ is the pose at $t{+}1$; for future-force
readout the predictor state is computed with that last action replaced by the
previous one, and the raw-history GRU receives the same history. Contact onset
marks $\|F\|>10$\,N within the next four steps while $\|F_t\|<10$\,N. The
future-force readout is a two-layer MLP (256--128 units) with early stopping,
fitted on training-task probe episodes.

\begin{table}[H]
\caption{\textbf{Flexiv, all metrics} (mean $\pm$ s.d.\ over three seeds;
IID: held-out episodes of training tasks; OOD: held-out tasks). Force and
position are read from the frame embedding; contact-onset AUC from the
predictor state. The untrained VFX network is evaluated at initialization.}
\label{tab:cfg1}
\begin{center}
\small
\resizebox{\linewidth}{!}{\begin{tabular}{llccccc}
\toprule
\rowcolor{tabhead} Model & Episodes & force IID & force OOD & position IID & position OOD & onset AUC OOD \\
\midrule
V & 800 & $-0.05\pm0.01$ & $-0.23\pm0.01$ & $0.19\pm0.04$ & $0.11\pm0.07$ & $0.68\pm0.01$ \\
 & 2,947 & $-0.06\pm0.01$ & $-0.28\pm0.13$ & $0.16\pm0.09$ & $0.03\pm0.12$ & $0.71\pm0.03$ \\
\midrule
VX & 800 & $-0.03\pm0.01$ & $-0.22\pm0.07$ & $0.31\pm0.03$ & $0.29\pm0.03$ & $0.71\pm0.04$ \\
 & 2,947 & $-0.04\pm0.01$ & $-0.21\pm0.02$ & $0.30\pm0.00$ & $0.23\pm0.02$ & $0.76\pm0.05$ \\
\midrule
VF & 800 & $0.15\pm0.02$ & $-0.08\pm0.05$ & $0.96\pm0.01$ & $0.96\pm0.01$ & $0.83\pm0.03$ \\
 & 2,947 & $0.14\pm0.04$ & $-0.16\pm0.08$ & $0.96\pm0.01$ & $0.96\pm0.01$ & $0.85\pm0.01$ \\
\midrule
\rowcolor{tabours} VFX & 800 & $0.93\pm0.00$ & $0.90\pm0.01$ & $0.97\pm0.00$ & $0.97\pm0.00$ & $0.90\pm0.01$ \\
\rowcolor{tabours}  & 2,947 & $0.92\pm0.01$ & $0.88\pm0.03$ & $0.97\pm0.00$ & $0.97\pm0.00$ & $0.88\pm0.00$ \\
\midrule
untrained VFX & none & $0.94$ & $0.91$ & $0.96$ & $0.98$ & $0.88$ \\
\bottomrule
\end{tabular}}
\end{center}
\end{table}

\paragraph{Data scale.} With $3.7\times$ more training episodes, future-force
readout of VFX rises from $0.40$ to $0.47$ on held-out tasks in every seed,
while VF stays below zero at both sizes (Table~\ref{tab:futfair}). The GRU
reference reads the same 16-step history as the models, and an MLP on the
current force and pose alone reaches $0.38$.

\begin{table}[H]
\caption{\textbf{Flexiv future force} ($0.4$\,s ahead; $R^2$, mean $\pm$ s.d.\
over three seeds), with IID and held-out-task (OOD) splits.}
\label{tab:futfair}
\begin{center}
\small
\begin{tabular}{lcccc}
\toprule
\rowcolor{tabhead} & \multicolumn{2}{c}{$800$ episodes} & \multicolumn{2}{c}{$2{,}947$ episodes} \\
\rowcolor{tabhead} & IID & OOD & IID & OOD \\
\midrule
V & $-0.01\pm0.01$ & $-0.10\pm0.04$ & $-0.03\pm0.02$ & $-0.15\pm0.07$ \\
VX & $0.07\pm0.03$ & $-0.07\pm0.04$ & $0.08\pm0.05$ & $-0.09\pm0.06$ \\
VF & $0.09\pm0.03$ & $-0.19\pm0.06$ & $0.09\pm0.02$ & $-0.10\pm0.05$ \\
\rowcolor{tabours} VFX & $0.39\pm0.01$ & $0.40\pm0.02$ & $0.43\pm0.01$ & $0.47\pm0.00$ \\
\midrule
untrained model, fused inputs & $0.37$ & $0.38$ & $0.37$ & $0.38$ \\
GRU on raw 16-step history & $0.44$ & $0.47$ & $0.44$ & $0.46$ \\
MLP on current force and pose & $0.40$ & $0.38$ & $0.40$ & $0.38$ \\
persistence & $0.23$ & $0.31$ & $0.23$ & $0.31$ \\
\bottomrule
\end{tabular}
\end{center}
\end{table}

\paragraph{KUKA.} On the KUKA configuration (single seed; clipping
thresholds computed over all episodes), only VFX reads force on held-out tasks
(Table~\ref{tab:kuka}).

\begin{table}[H]
\caption{\textbf{KUKA} (one seed). Force and position readout from the frame
embedding and contact-onset AUC (IID / held-out tasks).}
\label{tab:kuka}
\begin{center}
\small
\begin{tabular}{lccc}
\toprule
\rowcolor{tabhead} Model & force & position & onset AUC \\
\midrule
V & $-0.05$ / $-0.49$ & $0.76$ / $0.48$ & $0.70$ / $0.54$ \\
VX$_t$ & $-0.04$ / $-0.61$ & $0.77$ / $0.46$ & $0.80$ / $0.72$ \\
VX & $-0.02$ / $-0.70$ & $0.75$ / $0.50$ & $0.77$ / $0.69$ \\
\rowcolor{tabours} VFX & $0.87$ / $0.56$ & $0.98$ / $0.98$ & $0.86$ / $0.70$ \\
\bottomrule
\end{tabular}
\end{center}
\end{table}

\end{document}

%% file: math_commands.tex
\usepackage{amsmath,amsfonts,bm}

\def\eqref#1{equation~\ref{#1}}

\def\1{\bm{1}}

\DeclareMathAlphabet{\mathsfit}{\encodingdefault}{\sfdefault}{m}{sl}
\SetMathAlphabet{\mathsfit}{bold}{\encodingdefault}{\sfdefault}{bx}{n}

